\PassOptionsToPackage{table}{xcolor}
\documentclass{article}
\usepackage{iclr2027_conference,times}
\usepackage{preprint}
\usepackage[T1]{fontenc}
\usepackage{amsmath,amssymb}
\usepackage{booktabs}
\usepackage{graphicx}
\usepackage{float}
\usepackage[section,above]{placeins}
\usepackage{array}
\usepackage{tabularx}
\usepackage[hidelinks]{hyperref}
\usepackage{url}
\usepackage[table]{xcolor}
\usepackage{enumitem}
\usepackage{multirow}

\usepackage{amsthm}

\theoremstyle{definition}
\newtheorem{definition}{Definition}

\newcommand{\researchquestions}[2]{%
  \par\medskip
  \noindent
  \begingroup
  \setlength{\fboxsep}{7pt}%
  \fbox{%
    \parbox{\dimexpr\linewidth-2\fboxsep-2\fboxrule\relax}{%
      \textbf{RQ1.} #1\par\medskip
      \textbf{RQ2.} #2
    }%
  }%
  \endgroup
  \par\medskip
}

\newcommand{\answer}[1]{%
  \par\medskip
  \noindent
  \begingroup
  \setlength{\fboxsep}{7pt}%
  \fbox{%
    \parbox{\dimexpr\linewidth-2\fboxsep-2\fboxrule\relax}{%
      #1%
    }%
  }%
  \endgroup
  \par\medskip
}

\newcommand{\tablehead}[1]{\multicolumn{1}{c}{\bfseries\boldmath #1}}
\newcommand{\tableemph}[1]{{\bfseries\boldmath #1}}


\graphicspath{{figures/}}

\begin{document}

\title{\raggedright To Cache or Not to Cache: Understanding and Mitigating Safety Incompatibility in Cross-Model KV Cache Reuse}

\author{Zhiwei Lin$^{1}$ \quad Jiahao Liu$^{2}$ \quad Yan Liu$^{2}$ \quad Chengpeng Wang$^{1}$\\
{\normalfont $^{1}$National University of Singapore \quad $^{2}$Ant Group}\\
\texttt{zhiweil@u.nus.edu} \quad \texttt{ljiahaomail@gmail.com}\\
\texttt{bencao.ly@antgroup.com} \quad \texttt{wang-chengpeng@nus.edu.sg}}

\hypersetup{
  pdftitle={To Cache or Not to Cache: Understanding and Mitigating Safety Incompatibility in Cross-Model KV Cache Reuse},
  pdfauthor={Zhiwei Lin, Jiahao Liu, Yan Liu, Chengpeng Wang}
}

\maketitle

\begin{abstract}
Cross-model KV cache sharing reduces prefill computation by allowing a receiver model to reuse prefix KV states generated by a related sender model. Existing methods optimize such reuse for latency and generation quality, but its safety implications remain unclear. In this work, we systematically evaluate this issue across diverse sender--receiver pairs and reuse configurations. We refer to deviations from the receiver's native safety behavior as \emph{safety incompatibility}. Our study shows that both the direction and magnitude of safety incompatibility depend on the model pair and reuse configuration, with value reuse causing substantially larger deviations than key reuse across all evaluated pairs. Furthermore, we propose \textsc{VReparo}, a technique that learns per-layer and per-KV-head affine maps offline to align reused sender value tensors with their receiver-native counterparts before decoding.
Across two model pairs under full KV reuse and the state-of-the-art selective reuse setting,
\textsc{VReparo} reduces the average HarmBench attack
success rate (ASR) from 57.28\% to 39.21\% and the average StrongREJECT
score from 0.4215 to 0.2163 upon our benchmarks, while retaining an average Time to First Token (TTFT) reduction of 85.62\%.
\end{abstract}

\section{Introduction}
\label{sec:introduction}

KV caching and prefix reuse are important optimizations for efficient LLM serving.
Widely used serving systems such as vLLM~\citep{kwon2023pagedattention} manage
and reuse KV caches when requests share a common prefix. More recently, work
has extended KV cache reuse across related models: a \emph{sender model}
generates a prefix KV cache that is reused by a \emph{receiver model} to avoid
redundant prefill computation. For example, Activated LoRA allows its adapters to reuse the base model's cache for tokens preceding adapter activation~\citep{greenewald2025activated,li2025activated}, while
DroidSpeak enables KV cache sharing between related models through selective
layer recomputation for the shared prefix, while reusing sender KV
states at the remaining layers~\citep{liu2026droidspeak}.

A KV cache, however, is not merely a reusable storage artifact. Its model-generated key and value states directly participate in subsequent attention computations. Recent studies show that quantization of a model's own KV states can disrupt safety alignment even when conventional task-quality metrics remain largely preserved~\citep{xu2026alignment}, and that cache compression can also affect safety behavior~\citep{ma2026efficiencysafety}. Cross-model KV cache reuse introduces a distinct setting: the reused states are generated by a sender model and consumed by a different receiver model, rather than transformed from the receiver's own states.
Existing cross-model reuse techniques have primarily evaluated generation quality and serving performance~\citep{li2025activated, liu2026droidspeak}, leaving unclear whether reuse also preserves the receiver's native safety behavior.

\begin{figure}[t]
\centering
\includegraphics[width=.80\linewidth]{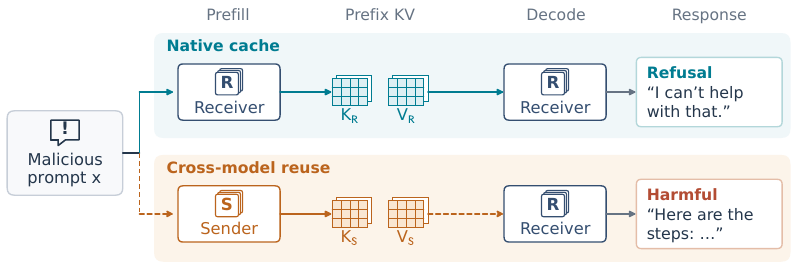}
\caption{One form of safety incompatibility in cross-model KV cache reuse,
where reusing sender-generated KV states changes a receiver-native safe
response into a harmful one.}
  \label{fig:study-event}
\end{figure}

In this work, we study whether cross-model KV cache reuse changes the safety behavior of the receiver. We refer to deviations from this receiver-native behavior as \emph{safety incompatibility}. Such incompatibility can manifest in different directions: reuse may turn a response that is safe under native inference into a harmful one, or may alter safety behavior in the opposite direction. Figure~\ref{fig:study-event} illustrates the former case, where the receiver refuses a malicious request under native inference but produces a harmful response when reusing a sender-generated KV cache.

We systematically investigate safety incompatibility across different model pairs and reuse configurations. Specifically, we study base/instruction-tuned model pairs from multiple model families and task-specific LoRI-D variants derived from the same base model~\citep{zhang2025lori}. We consider full-KV, K-only, and V-only reuse to examine how the reused cache components affect receiver safety. Our results show substantial safety incompatibility, whose magnitude and direction depend on both the model pair and the reuse configuration. For example, full-KV reuse from Llama-3-8B to Llama-3-8B-Instruct increases the HarmBench~\citep{mazeika2024harmbench} ASR from 6.37\% to 31.21\% and the average StrongREJECT~\citep{souly2024strongreject} score from 0.0094 to 0.2477. In contrast, for the Qwen3-8B base/instruction-tuned pair, full-KV reuse decreases the HarmBench ASR from 6.82\% to 4.55\%, while still changing the HarmBench safety labels on 3.79\% of the retained prompts.
The reuse configuration further affects the extent of this incompatibility. Across all evaluated model pairs on HarmBench, V-only reuse produces a larger deviation from the receiver-native ASR than K-only reuse. For the Llama-3-8B base/instruction-tuned pair, V-only and K-only reuse yield HarmBench ASRs of 57.25\% and 18.12\%, respectively. For the Qwen3-8B base/instruction-tuned pair, the corresponding ASRs are 9.30\% and 6.98\% under V-only and K-only reuse, respectively.

To mitigate this incompatibility, we propose \textsc{VReparo}, which learns affine maps offline from paired reused and receiver-native values. It applies these maps once to reused prefix values before receiver continuation, without constructing an online reference cache. The maps leave keys, recomputed values, and model weights unchanged.
Across four settings covering two model pairs under full and selective KV reuse, \textsc{VReparo} reduces the average HarmBench ASR from 57.28\% to 39.21\% and the average StrongREJECT score from 0.4215 to 0.2163. With resident sender caches, it retains an average TTFT reduction of 85.62\% relative to receiver-native inference.

\paragraph{Contributions.}
We make three main contributions in this work.
\begin{itemize}[leftmargin=*]
\item We formulate the problem of \emph{safety incompatibility} in cross-model
KV cache reuse as deviations from receiver-native safety behavior
caused by reusing sender-generated KV caches.

\item We systematically characterize how cross-model KV cache reuse
alters receiver-native safety behavior across model pairs and reuse
configurations, identifying the dependence of these changes on reuse
direction and key/value components.

\item We propose and evaluate \textsc{VReparo}, a lightweight technique that mitigates safety incompatibility while preserving cache-reuse efficiency, and quantify its effects on safety, utility, and runtime overhead.
\end{itemize}

\section{Related Work}
\label{sec:related-work}

Existing work on KV caches has largely focused on improving serving efficiency.
KV cache optimization spans serving-system memory management and prefix
sharing~\citep{kwon2023pagedattention}, token
retention~\citep{zhang2023h2o,li2024snapkv},
quantization~\citep{liu2024kivi}, and cross-layer
sharing~\citep{yang2024kvsharer}. Recent methods extend reuse across model
weights through cache reuse for tokens preceding adapter activation~\citep{greenewald2025activated,li2025activated} and selective layer recomputation~\citep{liu2026droidspeak}. Other approaches use closed-form affine maps~\citep{heo2026crossmodel}, head-local mappings with attention-aligned calibration~\citep{qu2026cachebridge}, reconstruction from compact semantic codes with selective patching~\citep{ma2026semanticcache}, or learned cache fusion for communication between models~\citep{fu2026cachetocache}, primarily targeting
efficiency and generation quality.

Alongside these efficiency advances, recent work has begun to examine the
safety and security implications of modifying or reusing KV caches.
Safety studies show that compression can affect safety-sensitive
capabilities~\citep{liu2025fundamental} and either weaken or strengthen
jailbreak attacks depending on the compression regime~\citep{ma2026efficiencysafety}.
Low-bit KV quantization can also degrade refusal behavior while conventional
utility remains largely stable~\citep{xu2026alignment}. Beyond compression,
HijackKV exploits position-independent reuse to inject attacker-controlled
context through cached states associated with benign text~\citep{zhang2026hijackkv}.
Our study instead examines changes in safety behavior when reusing caches
produced by related models, without adversarial cache construction.

These risks have motivated cache-level mitigations.
Existing mitigations include eviction policies that preserve instructions
and reduce system-prompt leakage~\citep{chen2025pitfalls}, history-aware
cache merging in Safe-CAM~\citep{ma2026efficiencysafety}, quantization repair
guided by per-channel reduction~\citep{xu2026alignment},
and refusal-aware token retention in
AnchorKV~\citep{ni2026anchorkv}. These methods primarily address transformations
of a model's own cache. We study safety incompatibility across
sender--receiver pairs and reuse configurations, and develop
\textsc{VReparo} to map reused values toward receiver-native values
while retaining the efficiency benefits of cross-model reuse.

\section{Background and Problem Statement}
\label{sec:problem}

\subsection{Preliminaries of Cross-Model KV Cache Reuse}
\label{sec:kv-reuse}

\paragraph{KV cache in LLM inference.}
Transformer-based LLM inference consists of a \emph{prefill phase} and a
\emph{decode phase}. During prefill, the model processes the input context and
stores the key (K) and value (V) tensors produced at each Transformer layer.
During autoregressive decode, each newly generated token attends to these cached
states and appends its own K and V tensors to the cache, avoiding recomputation
over previously processed tokens.
For a model $M$ with $L_M$ transformer layers and a token sequence $z$, we
write its KV cache as
\begin{equation}
  C_M(z)
  =
  \left(
    \left(
      K_M^\ell(z),
      V_M^\ell(z)
    \right)
  \right)_{\ell=1}^{L_M},
  \label{eq:kv-cache}
\end{equation}
where $K_M^\ell(z)$ and $V_M^\ell(z)$ denote the cached K and V tensors at
layer $\ell$. These tensors are determined by the hidden states produced during
the corresponding forward execution. Consequently, changing the cached states
used at one layer can affect the receiver-computed states at subsequent layers.

\paragraph{Cross-model KV cache reuse.}
Cross-model KV cache reuse uses KV tensors produced by a \emph{sender model}
$S$ to reduce the prefill computation performed by a \emph{receiver model}
$R$. Let $\mathcal L$ denote the set of layer positions at which sender KV
tensors may be reused by the receiver.
To characterize which components are reused, we define the
\emph{cross-model reuse configuration}
\begin{equation}
  \rho=(\mathcal U_K,\mathcal U_V),
  \qquad
  \mathcal U_K,\mathcal U_V\subseteq\mathcal L,
  \label{eq:reuse-configuration}
\end{equation}
where $\mathcal U_K$ and $\mathcal U_V$ specify the layers that reuse sender K
and V tensors, respectively. Components not selected for reuse are
supplied by the receiver. In the diagnostic experiments in
Section~\ref{sec:characterization}, sender and receiver-native caches are
constructed independently from the same token sequence, and the selected
components are combined before receiver decoding.
K-only uses sender keys and receiver-native values.
V-only uses receiver-native keys and sender values.
We use DroidSpeak~\citep{liu2026droidspeak} to evaluate \textsc{VReparo}.
For the shared prefix, it recomputes KV states in a selected receiver layer
block from the sender's intermediate hidden states and reuses sender KV
states at all other layers. The receiver then processes the prompt suffix.

\paragraph{LLM inference under reuse configurations.}
Let the tokenized prompt be $x=(x_1,\ldots,x_T)$, with
$x_{<T}=(x_1,\ldots,x_{T-1})$. All configurations initialize receiver
continuation with a cache for $x_{<T}$ and process the same boundary token $x_T$.
We write $C_{S\rightarrow R}^{\rho}(z)$ for the cache constructed for $z$
under configuration $\rho$, using the cache-construction procedure specified
for the experiment. This construction may combine a reused shared
prefix with a suffix processed by the receiver.
Let $\operatorname{Infer}_R(x_T,C)$ denote receiver inference initialized with
cache $C$, which processes $x_T$ and continues autoregressive generation.
Under cross-model reuse,
\begin{equation}
  y_{S\rightarrow R}^{\rho}(x)
  =
  \operatorname{Infer}_R
  \left(
    x_T,
    C_{S\rightarrow R}^{\rho}(x_{<T})
  \right).
  \label{eq:reuse-inference}
\end{equation}
When no cross-model reuse is applied, the receiver constructs the prefix cache
itself and performs \emph{receiver-native inference}, whose output is denoted by $y_R^{\mathrm{native}}(x)$.

\subsection{Safety Incompatibility}
\label{sec:safety-incompatibility}

Because KV caches encode model-dependent intermediate states, reusing them can
alter not only the receiver's task behavior but also its safety behavior. We
refer to deviations from the receiver's native safety behavior as
\emph{safety incompatibility}. 
We quantify safety incompatibility through the direction and magnitude
of changes in the receiver's safety scores.

\begin{definition}[Safety Score Shift]
Given a prompt set $D$, let $\sigma(y,x)\in[0,1]$ denote a harmfulness evaluator
for response $y$ to prompt $x$, where $0$ indicates a safe response and larger
values indicate greater safety risk. The \emph{safety score shift} ($\Delta S$)
of reuse configuration $\rho$ for sender $S$ and receiver $R$ over $D$ is
\begin{equation}
  \Delta S_{S\rightarrow R}^{\rho}(D)
  =
  \frac{1}{|D|}
  \sum_{x\in D}
  \left[
    \sigma\!\left(y_{S\rightarrow R}^{\rho}(x),x\right)
    -
    \sigma\!\left(y_R^{\mathrm{native}}(x),x\right)
  \right].
  \label{eq:safety-score-shift}
\end{equation}
\end{definition}

\begin{definition}[Safety Degradation and Improvement]
Given the prompt set $D$ and harmfulness evaluator $\sigma$, the
\emph{safety degradation} (SD) and \emph{safety improvement} (SI) are
\begin{equation}
\begin{aligned}
  \mathrm{SD}_{S\rightarrow R}^{\rho}(D)
  &=
  \frac{1}{|D|}
  \sum_{x\in D}
  \max\left\{
    \sigma\!\left(y_{S\rightarrow R}^{\rho}(x),x\right)
    -
    \sigma\!\left(y_R^{\mathrm{native}}(x),x\right),
    0
  \right\}, \\
  \mathrm{SI}_{S\rightarrow R}^{\rho}(D)
  &=
  \frac{1}{|D|}
  \sum_{x\in D}
  \max\left\{
    \sigma\!\left(y_R^{\mathrm{native}}(x),x\right)
    -
    \sigma\!\left(y_{S\rightarrow R}^{\rho}(x),x\right),
    0
  \right\}.
\end{aligned}
\label{eq:safety-degradation-improvement}
\end{equation}
\end{definition}

\begin{definition}[Safety Deviation]
The \emph{safety deviation} (SDV) measures the overall magnitude of the
deviation from the receiver's native safety behavior:
\begin{equation}
  \mathrm{SDV}_{S\rightarrow R}^{\rho}(D)
  =
  \frac{1}{|D|}
  \sum_{x\in D}
  \left|
    \sigma\!\left(y_{S\rightarrow R}^{\rho}(x),x\right)
    -
    \sigma\!\left(y_R^{\mathrm{native}}(x),x\right)
  \right|.
  \label{eq:safety-deviation}
\end{equation}
\end{definition}

These metrics capture complementary aspects of safety incompatibility across model pairs and
reuse configurations.
Concretely, a smaller SDV indicates greater safety compatibility, with
$\mathrm{SDV}=0$ indicating that reuse preserves the receiver's native safety
scores for all prompts in $D$. Conversely, a larger SDV indicates greater
safety incompatibility. A small $|\Delta S|$ alone does not necessarily imply
safety compatibility, since SD and SI may offset each
other even when their combined magnitude is large.
We use this formulation to investigate two research questions:

\researchquestions
{Does cross-model KV-cache reuse introduce safety incompatibility of the receiver model?}
{Which factors can affect safety incompatibility of the receiver model?}

In Section~\ref{sec:characterization}, we use these metrics to examine whether cross-model KV cache reuse
preserves receiver-native safety behavior and how the observed incompatibility
varies across reuse settings.

\section{Characterizing Safety Incompatibility}
\label{sec:characterization}

\subsection{Study Methodology}
\label{sec:characterization-methodology}

To answer RQ1, we consider two types of model pairs,
following the evaluation setting used in DroidSpeak~\citep{liu2026droidspeak}. 
The first type pairs base models with their instruction-tuned counterparts from several model families, allowing us to examine whether the safety effect of KV
cache reuse generalizes across models. The second type consists of
task-specific LoRI-D variants that share the same base model and major
LoRA configuration, providing a more controlled comparison across
training objectives. For each model pair, we fix the receiver and tokenized prompt and compare full KV reuse with the receiver's native inference. 

To answer RQ2, we use the same sender--receiver pairs as in RQ1.
To examine the effect of cache composition, we compare full KV, K-only, and
V-only reuse with receiver-native inference. Following the cache construction described in
Section~\ref{sec:kv-reuse}, components not reused from the sender
are taken from the receiver-native cache for the same tokenized prompt.

\begin{table*}[t]
\centering
\caption{Safety incompatibility under full KV cache reuse. HarmBench $\Delta S$ is in percentage points (pp) and SD/SI/SDV in \%. StrongREJECT metrics use the original score scale.}
\label{tab:rq1-full-reuse}
\setlength{\tabcolsep}{3.1pt}
\renewcommand{\arraystretch}{1.12}
\begin{tabular*}{\linewidth}{@{\extracolsep{\fill}}lc*{8}{r}@{}}
\toprule
\tablehead{\multirow[c]{2}{*}{Model}}
& \tablehead{\multirow[c]{2}{*}{(Sender, Receiver)}}
& \multicolumn{4}{c}{\textbf{HarmBench}}
& \multicolumn{4}{c}{\textbf{StrongREJECT}} \\
\cmidrule(lr){3-6}\cmidrule(lr){7-10}
& &
\tablehead{$\Delta S$}
& \tablehead{SD}
& \tablehead{SI}
& \tablehead{SDV}
& \tablehead{$\Delta S$}
& \tablehead{SD}
& \tablehead{SI}
& \tablehead{SDV} \\
\midrule
Qwen3-4B & (Base, Instruct) & +10.13 & 13.29 & 3.16 & 16.46 & +.0495 & .0716 & .0221 & .0938 \\
Qwen3-8B & (Base, Instruct) & -2.27 & 0.76 & 3.03 & 3.79 & -.0260 & .0077 & .0337 & .0414 \\
Llama-3.2-3B & (Base, Instruct) & +8.67 & 13.33 & 4.67 & 18.00 & +.0781 & .1009 & .0227 & .1236 \\
Llama-3-8B & (Base, Instruct) & +24.84 & 28.03 & 3.18 & 31.21 & +.2383 & .2387 & .0004 & .2391 \\
Gemma-2-2B & (Base, Instruct) & +20.25 & 20.89 & 0.63 & 21.52 & +.1564 & .1576 & .0012 & .1589 \\
Mistral-7B-v0.3 & (Base, Instruct) & +9.43 & 16.98 & 7.55 & 24.53 & +.0882 & .1594 & .0712 & .2306 \\
\midrule
LoRI-D & (Code, Safety) & +54.81 & 57.78 & 2.96 & 60.74 & +.5648 & .5719 & .0071 & .5790 \\
LoRI-D & (Safety, Code) & -44.53 & 3.65 & 48.18 & 51.82 & -.4512 & .0187 & .4699 & .4886 \\
\bottomrule
\end{tabular*}
\end{table*}

\subsection{Experimental Subjects and Settings}
\label{sec:characterization-setup}

For RQ1, we evaluate six base/instruction-tuned model pairs
from four families, which include Qwen3-4B, Qwen3-8B, Llama-3.2-3B,
Llama-3-8B, Gemma-2-2B, and Mistral-7B-v0.3.
The base model serves as the sender and the instruction-tuned model as the receiver.
We also study the Code and Safety variants of LoRI-D~\citep{zhang2025lori},
which share Meta-Llama-3-8B and the same major adapter configuration.
The Code and Safety variants are fine-tuned on CodeAlpaca and SaferPaca,
respectively. We evaluate the Code/Safety pair in both directions.
RQ2 uses the same model pairs to compare different K/V reuse configurations.

Following the benchmark choices in prior LLM safety
studies~\citep{yao2025mousetrap,chen2026metacipher},
we use HarmBench~\citep{mazeika2024harmbench} and
StrongREJECT~\citep{souly2024strongreject} as the main evaluation benchmarks
for RQ1, and HarmBench for the K/V separation experiments in RQ2.
We evaluate on 159 harmful requests from HarmBench and 313 from StrongREJECT before output validity filtering.
The benchmarks provide complementary assessments: HarmBench evaluates
whether a response exhibits the target harmful behavior, whereas
StrongREJECT measures the extent to which a response provides harmful
assistance.
Notably, we retain only prompts whose outputs pass the same
validity check in all compared conditions and report the retained sample size.
Implementation details and output validity checks are provided in
Appendix~\ref{app:characterization-settings}.

We instantiate $\sigma$ with binary HarmBench judgments in $\{0,1\}$
and StrongREJECT scores in $[0,1]$ to compute $\Delta S$, SD, SI, and SDV
as defined in Section~\ref{sec:safety-incompatibility}. We also report HarmBench ASR,
the percentage of responses judged to exhibit the requested harmful behavior,
and the average StrongREJECT score.
HarmBench SD, SI, and SDV are percentages of prompts with label
transitions, whereas $\Delta S$ is an ASR difference in percentage points (pp).
Lower average HarmBench ASRs and StrongREJECT scores
indicate safer responses, whereas lower SDV indicates greater agreement
with receiver-native safety scores on the retained prompts.

\subsection{Empirical Findings}

\subsubsection{Safety Incompatibility under Cross-Model KV-Cache Reuse}
\label{sec:eval-rq1}

\paragraph{Results.}
As shown in Table~\ref{tab:rq1-full-reuse}, full KV cache reuse does not
consistently preserve receiver-native safety behavior, even among
responses that pass the validity check. 
When the base model supplies the cache to its
instruction-tuned counterpart, harmfulness increases for five of the
six model pairs under both evaluators. Qwen3-8B is the exception in 
both HarmBench ASR and average StrongREJECT score, but reuse
still changes the HarmBench label on 3.79\% of retained prompts. Across the six base/instruction-tuned model pairs, HarmBench SDV ranges from 3.79\% to 31.21\%.
As shown in Table~\ref{tab:rq1-harmbench-detail} in Appendix~\ref{app:rq1-details}, full reuse raises HarmBench ASR for Gemma-2-2B from 1.27\% to 21.52\%.
For Mistral-7B-v0.3, the ASR increase is 9.43 percentage points, while 24.53\% of the
HarmBench labels change, showing that changes in ASR do not capture
all safety changes.

The LoRI-D variants share the same base model, providing a more
controlled comparison. Among the evaluated pairs, the largest increase
in harmfulness occurs when the LoRI-D Safety
variant reuses the KV cache from the Code variant. As shown in Tables~\ref{tab:rq1-harmbench-detail} and~\ref{tab:rq1-strongreject-detail}, full reuse raises
HarmBench ASR from 25.93\% to 80.74\% and the average StrongREJECT score from
.1093 to .6741. In the reverse direction, the Code variant reuses
the KV cache from the Safety variant. Both ASR and the average StrongREJECT score decrease,
but HarmBench SDV is 51.82\%, indicating that safety judgments
change for more than half of the retained prompts.
Despite the lower ASR, HarmBench SD is 3.65\%, with
responses to 5 of the 137 retained prompts changing from
being judged safe to harmful.
Thus, sharing the same foundation model does
not ensure safety compatibility.

\answer{
\noindent\textbf{Answer to RQ1.}
Cross-model KV cache reuse introduces safety incompatibility across all
evaluated pairs. SD and SI can coexist. Several responses become more harmful
even when average harmfulness over the evaluated prompts decreases.
}

\subsubsection{Factors Affecting Safety Incompatibility}
\label{sec:eval-rq2}
\label{sec:rq2-components}
\begin{table}[t]
\centering
\caption{HarmBench ASRs (\%). $\Delta_{\mathrm{K}-\mathrm{N}}$,
$\Delta_{\mathrm{V}-\mathrm{N}}$, and $\Delta_{\mathrm{KV}-\mathrm{N}}$ are K-only,
V-only, and Full KV ASRs minus Native ASR, respectively, in pp.
Positive differences indicate increased harmfulness.}
\label{tab:rq2-kv-components}
\setlength{\tabcolsep}{3.5pt}
\renewcommand{\arraystretch}{1.12}
\begin{tabular*}{\linewidth}{@{\extracolsep{\fill}}l*{8}{r}@{}}
\toprule
\tablehead{\multirow[c]{2}{*}{Model / Pair}}
& \tablehead{\multirow[c]{2}{*}{$n$}}
& \multicolumn{4}{c}{\textbf{ASR (\%)}}
& \multicolumn{3}{c}{\textbf{Change from Native (pp)}} \\
\cmidrule(lr){3-6}\cmidrule(lr){7-9}
& &
\tablehead{Native}
& \tablehead{K-only}
& \tablehead{V-only}
& \tablehead{Full KV}
& \tablehead{$\Delta_{\mathrm{K}-\mathrm{N}}$}
& \tablehead{$\Delta_{\mathrm{V}-\mathrm{N}}$}
& \tablehead{$\Delta_{\mathrm{KV}-\mathrm{N}}$} \\
\midrule
\multicolumn{9}{@{}l}{\emph{(Base, Instruct)}} \\
Qwen3-4B & 156 & 10.90 & 13.46 & 25.64 & 21.15 & +2.56 & +14.74 & +10.26 \\
Qwen3-8B & 129 & 6.98 & 6.98 & 9.30 & 4.65 & 0.00 & +2.33 & -2.33 \\
Llama-3.2-3B & 147 & 8.84 & 7.48 & 28.57 & 17.01 & -1.36 & +19.73 & +8.16 \\
Llama-3-8B & 138 & 5.80 & 18.12 & 57.25 & 28.99 & +12.32 & +51.45 & +23.19 \\
Gemma-2-2B & 121 & 1.65 & 3.31 & 35.54 & 23.97 & +1.65 & +33.88 & +22.31 \\
Mistral-7B-v0.3 & 158 & 58.23 & 52.53 & 70.89 & 67.72 & -5.70 & +12.66 & +9.49 \\
\midrule
\multicolumn{9}{@{}l}{\emph{LoRI-D}} \\
(Code, Safety) & 127 & 26.77 & 24.41 & 81.89 & 81.10 & -2.36 & +55.12 & +54.33 \\
(Safety, Code) & 128 & 82.03 & 82.81 & 40.62 & 37.50 & +0.78 & -41.41 & -44.53 \\
\bottomrule
\end{tabular*}
\end{table}

\noindent\textbf{Reuse configuration.}
We compare K-only, V-only, and full KV reuse to examine how the reused
cache components affect receiver safety.
In Table~\ref{tab:rq2-kv-components}, $n$ counts prompts passing the
validity check under all four conditions, out of 159.
As shown in Table~\ref{tab:rq2-kv-components}, ASR differs
more from the receiver-native baseline under V-only reuse than
under K-only reuse across all eight sender--receiver pairs.
For the six base/instruction-tuned model pairs, V-only reuse
has a higher ASR than K-only reuse.
Full KV reuse has a lower ASR than V-only reuse across all eight pairs.
For the Qwen3-8B base/instruction-tuned pair, V-only reuse raises ASR from 6.98\% to
9.30\%, whereas full KV reuse lowers it to 4.65\%. Thus, the effect of
V-only reuse does not fully describe the effect of reusing both
keys and values.

\noindent\textbf{Model pair.}
The magnitude and direction of the changes also vary across model pairs.
For example, V-only reuse increases ASR by 2.33 percentage points for
Qwen3-8B and 51.45 percentage points for Llama-3-8B. For the LoRI-D variants, when Safety is
the receiver and Code is the sender, V-only reuse raises ASR from 26.77\%
to 81.89\%, close to 81.10\% under full KV reuse. K-only reuse yields an ASR
of 24.41\%, close to the native value. In the reverse direction, when Code
reuses the cache from Safety, V-only reuse lowers ASR from 82.03\% to 40.62\%,
while K-only reuse yields 82.81\%. Thus, the same reuse configuration can
increase or decrease harmfulness across model pairs.

\answer{
\noindent\textbf{Answer to RQ2.}
Compared with receiver-native inference, V-only reuse leads to larger
changes in ASR than K-only reuse across all evaluated pairs on HarmBench.
Whether reuse increases or decreases ASR depends on the model pair and reuse
configuration.
}

\section{Mitigating Safety Incompatibility}
\label{sec:approach}

\begin{figure}[t]
  \centering
  \includegraphics[width=\linewidth]{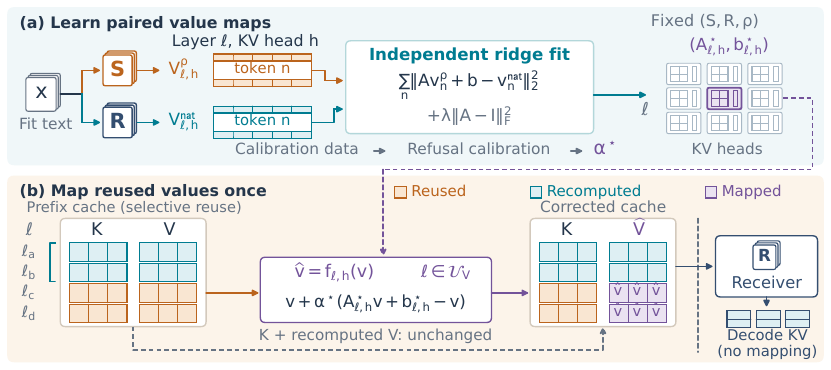}
  \caption{The workflow of \textsc{VReparo}. (a) Matching token positions at each
  layer and KV head provide paired reused and receiver-native values for
  affine fitting. Separate behavioral calibration selects $\alpha^\star$.
  (b) The fixed maps transform reused prefix values once. Keys,
  recomputed values, and newly generated KV states are not transformed.
  The layer layout illustrates selective reuse. Under full KV reuse,
  prefix values at all layers are mapped.}
  \label{fig:value-sanitization}
\end{figure}

Section~\ref{sec:eval-rq2} motivates correcting reused values,
although keys also affect safety. \textsc{VReparo} learns maps from reused
to receiver-native values offline, avoiding reference prefill for each
request. For each sender $S$, receiver $R$, and fixed reuse configuration
$\rho$, we fit the maps separately and apply them before decoding to reduce
safety deviation while retaining prefill savings.
Figure~\ref{fig:value-sanitization} illustrates both offline map fitting
and online value correction.

\paragraph{Data collection.}
We fit the affine maps without task or safety labels and
select the correction strength $\alpha$ using refusal rates on
a separate set of safe and unsafe prompts. For each fitting example, we use identical
tokenized inputs and prompt formatting to construct the cache under
$\rho$ and the receiver-native cache. We collect value vectors at matching
token positions in the same layer and KV head. For a layer whose values
are reused, let $v_n^{\rho}\in\mathbb{R}^{d_v}$ be a reused value vector and
$v_n^{\mathrm{nat}}\in\mathbb{R}^{d_v}$ the corresponding receiver-native
value vector, where $d_v$ is the head dimension. 
The index $n$ runs over selected prefill positions across fitting examples.
Layer and head indices are omitted for clarity.

\paragraph{Offline fitting.}
For each layer and KV head whose values are reused, we fit an affine map
with matrix $A\in\mathbb{R}^{d_v\times d_v}$ and bias
$b\in\mathbb{R}^{d_v}$:
\begin{equation}
    (A^\star,b^\star)=\arg\min_{A,b}
    \sum_{n=1}^{N}
    \left\|A v_n^{\rho}+b-v_n^{\mathrm{nat}}\right\|_2^2
    +\lambda\left\|A-I\right\|_F^2,
    \label{eq:value-map-fit}
\end{equation}
where $N$ is the number of collected pairs and $I$ is the identity matrix.
The regularization term encourages $A$ to remain close to the identity.
The bias is not penalized. This is a closed-form ridge fit and does not
update either language model. 
We use a full matrix for $A$ and fix the ridge coefficient $\lambda$ across
all evaluated model pairs and reuse configurations.

The full mapping may also change responses to benign prompts.
We therefore select a scalar strength $\alpha^*\in[0,1]$ on behavioral
calibration data, maximizing refusal on unsafe prompts subject to limits
on refusal of benign prompts. The strength is fixed across requests
in each $(S,R,\rho)$ setting.

\paragraph{Inference.}
Before receiver decoding, each reused value vector is replaced by
\begin{equation}
    \widehat v_n
    =v_n^{\rho}
    +\alpha^*\left(A^\star v_n^{\rho}+b^\star-v_n^{\rho}\right).
    \label{eq:value-map-inference}
\end{equation}
Thus $\alpha^*=0$ leaves the value unchanged, while $\alpha^*=1$ applies
the full fitted mapping. The mapping preserves cache dimensions and
leaves the key tensors unchanged. For full KV reuse, it is applied to
the sender's prompt values before the receiver processes the prompt
boundary token. For DroidSpeak, it is applied to reused prefix values
outside the receiver's recomputed layer block. The receiver then
processes the prompt suffix using the corrected prefix cache. Values
in the recomputed block are not mapped. In both cases, the receiver
continues autoregressive decoding with its original weights and
language-model head. The mapping is applied once to the reused prompt
cache. Values produced during subsequent decoding are not mapped.
No safety label is required at inference.

\section{Evaluation}
\label{sec:mitigation-evaluation}

We evaluate \textsc{VReparo}'s safety compatibility, question-answering
(QA) quality, benign refusal, and runtime under full KV reuse and selective
reuse with DroidSpeak~\citep{liu2026droidspeak}.

\subsection{Experimental Setup}
\label{sec:mitigation-eval-setup}

\paragraph{Evaluation settings.}
We evaluate two sender--receiver pairs under both reuse
configurations. For the LoRI-D pair, the Safety variant reuses
the KV cache from the Code variant. For the Llama-3-8B
base/instruction-tuned pair, the instruction-tuned model reuses
the KV cache from the base model.
Each setting compares receiver-native inference, uncorrected
reuse, and reuse with \textsc{VReparo}.
For DroidSpeak, we select the receiver layer block
to recompute based on QA performance.

\paragraph{Fitting and calibration.}
We fit separate affine maps for each sender--receiver pair and reuse
configuration using 32 HotpotQA~\citep{yang2018hotpotqa} and 32
HEx-PHI~\citep{qi2023finetuning} examples, without task or safety labels.
We select the correction strength $\alpha$ using string-based
refusal rates on XSTest~\citep{rottger2024xstest}.
We repeat this procedure with five random seeds for data sampling. The fitted maps
and selected strength remain fixed during evaluation within each run.

\paragraph{Benchmarks and reporting.}
Safety evaluation uses 127 HarmBench prompts after excluding a separate
32-prompt development subset, and all 313 StrongREJECT prompts. The development
subset is distinct from the fitting examples. Sampling details are provided in
Appendix~\ref{app:mitigation-settings}. We report
HarmBench ASR and average StrongREJECT score (0--1).
For both benchmarks, we also report $\Delta S$, SD, SI, and SDV
as defined in Section~\ref{sec:safety-incompatibility}, computed from paired per-prompt
scores relative to Native within the same configuration.
We evaluate QA quality using token-level F1, the harmonic mean
of token precision and recall.
Benign refusal is evaluated on 200 OR-Bench prompts~\citep{cui2024orbench}.
Full refusal means declining assistance without a substantive completion.

All conditions within each setting use the same fixed prompts,
without the response-based filtering used in Section~\ref{sec:characterization}.
These datasets were examined during earlier development.
Layer selection, fitting, calibration, and evaluation details are
provided in Appendix~\ref{app:mitigation-settings}.
Tables~\ref{tab:value-safety} and~\ref{tab:value-runtime} report safety
and latency. Table~\ref{tab:value-utility} in Appendix~\ref{app:mitigation-quality}
reports QA quality.

\subsection{\texorpdfstring{Experimental Results}{Experimental Results}}
\label{sec:eval-value-results}

\begin{table}[t]
\caption{Native-referenced safety metrics.
\textsc{VReparo} results are averaged over five runs.
HarmBench $\Delta S$ is in pp and SD/SI/SDV in \%.
StrongREJECT uses the original score scale.
Native metrics are zero. Bold marks the lower SDV per configuration.}
\label{tab:value-safety}
\label{tab:value-full-kv}
\label{tab:value-droidspeak}
\centering
\setlength{\tabcolsep}{2.4pt}
\renewcommand{\arraystretch}{1.12}
\begin{tabular*}{\linewidth}{@{\extracolsep{\fill}}lll*{8}{r}@{}}
\toprule
\tablehead{\multirow[c]{2}{*}{Model}}
& \tablehead{\multirow[c]{2}{*}{Config.}}
& \tablehead{\multirow[c]{2}{*}{Condition}}
& \multicolumn{4}{c}{\textbf{HarmBench}}
& \multicolumn{4}{c}{\textbf{StrongREJECT}} \\
\cmidrule(lr){4-7}\cmidrule(l){8-11}
& & &
\tablehead{$\Delta S$}
& \tablehead{SD}
& \tablehead{SI}
& \tablehead{SDV}
& \tablehead{$\Delta S$}
& \tablehead{SD}
& \tablehead{SI}
& \tablehead{SDV} \\
\midrule
LoRI-D & Full KV & Reuse & 50.39 & 54.33 & 3.94 & 58.27 & 0.5212 & 0.5337 & 0.0125 & 0.5462 \\
 & & \textsc{VReparo} & 34.96 & 38.90 & 3.94 & \tableemph{42.83} & 0.2746 & 0.2940 & 0.0194 & \tableemph{0.3134} \\
\addlinespace[4pt]
 & DroidSpeak & Reuse & 53.54 & 56.69 & 3.15 & 59.84 & 0.5140 & 0.5205 & 0.0065 & 0.5270 \\
 & & \textsc{VReparo} & 31.34 & 34.96 & 3.62 & \tableemph{38.58} & 0.2703 & 0.2846 & 0.0143 & \tableemph{0.2989} \\
\midrule
Llama-3-8B & Full KV & Reuse & 27.56 & 30.71 & 3.15 & 33.86 & 0.2398 & 0.2416 & 0.0018 & 0.2434 \\
 & & \textsc{VReparo} & 7.56 & 10.08 & 2.52 & \tableemph{12.60} & 0.0266 & 0.0291 & 0.0025 & \tableemph{0.0316} \\
\addlinespace[4pt]
 & DroidSpeak & Reuse & 25.20 & 25.98 & 0.79 & 26.77 & 0.1462 & 0.1466 & 0.0004 & 0.1470 \\
 & & \textsc{VReparo} & 10.55 & 11.97 & 1.42 & \tableemph{13.39} & 0.0291 & 0.0326 & 0.0036 & \tableemph{0.0362} \\
\bottomrule
\end{tabular*}
\end{table}

\paragraph{Safety and benign refusal.}
\label{sec:eval-value-full-kv}
\label{sec:eval-value-droidspeak}
Table~\ref{tab:value-safety} shows lower $\Delta S$, SD, and SDV
under \textsc{VReparo} on both benchmarks in all four settings.
For LoRI-D full reuse, HarmBench SDV falls from 58.27\% to 42.83\%,
and StrongREJECT SDV from 0.5462 to 0.3134.
Under Llama-3-8B DroidSpeak, HarmBench SDV falls from 26.77\% to 13.39\%
while the recomputation block remains fixed. This improvement comes from
mapping reused values outside that block, without enlarging it.
Table~\ref{tab:value-absolute} in Appendix~\ref{app:mitigation-absolute}
reports HarmBench ASR, average StrongREJECT scores, and OR-Bench full refusal rates.
HarmBench ASR and average StrongREJECT score decrease in every setting
but remain above Native. OR-Bench full refusal
also rises from 4.0\% to 12.6\% for LoRI-D full reuse, compared
with 11.0\% for Native.
Thus, the XSTest calibration constraint does not ensure preservation of benign refusal rates on OR-Bench.
SDV decreases in all five runs for every setting on both benchmarks.
Appendix~\ref{app:fitting-subsets} reports individual results.

Llama-3-8B's mean HotpotQA F1 changes from 43.43 to 43.42 with
DroidSpeak and from 9.68 to 10.55 with full reuse, below Native's 47.91.
Appendix~\ref{app:fitting-subsets} reports individual results on 248 HotpotQA prompts.
Value mapping does not generally restore native QA quality.

\paragraph{Runtime.}
\label{sec:eval-value-runtime}

On an NVIDIA A40, we time 50 fixed HotpotQA inputs using Hugging Face.
Native TTFT includes full receiver prefill. Both reuse conditions start from resident sender caches and, for DroidSpeak, intermediate states.
TTFT includes value mapping, receiver recomputation, and suffix processing where applicable.
We exclude sender prefill, cache transfer, tokenization, and distributed scheduling.
Appendix~\ref{app:mitigation-timing} details the full protocol.

\begin{table}[t]
\centering
\caption{Mean latency with resident sender caches. Speedup is Native TTFT
divided by \textsc{VReparo} TTFT.
Per-token generation time excludes one-time cache preparation.}
\label{tab:value-runtime}
\setlength{\tabcolsep}{3.5pt}
\renewcommand{\arraystretch}{1.12}
\begin{tabular*}{\linewidth}{@{\extracolsep{\fill}}ll*{4}{r}*{2}{c}@{}}
\toprule
\tablehead{\multirow[c]{2}{*}{Model}}
& \tablehead{\multirow[c]{2}{*}{Configuration}}
& \multicolumn{3}{c}{\textbf{TTFT (ms)}}
& \tablehead{\multirow[c]{2}{*}{Speedup}}
& \multicolumn{2}{c}{\textbf{Generation (ms/token)}} \\
\cmidrule(lr){3-5}\cmidrule(lr){7-8}
& &
\tablehead{Native}
& \tablehead{Reuse}
& \tablehead{\textsc{VReparo}}
&
&
\tablehead{Reuse}
& \tablehead{\textsc{VReparo}} \\
\midrule
LoRI-D & Full KV & 1,355.84 & 54.71 & 88.98 & \tableemph{$15.24\times$} & 51.15 & 51.15 \\
 & DroidSpeak & 1,344.16 & 201.69 & 299.11 & \tableemph{$4.49\times$} & 51.15 & 51.14 \\
\midrule
Llama-3-8B & Full KV & 1,350.37 & 54.62 & 88.80 & \tableemph{$15.21\times$} & 51.08 & 51.07 \\
 & DroidSpeak & 1,347.80 & 201.59 & 298.40 & \tableemph{$4.52\times$} & 51.04 & 51.03 \\
\bottomrule
\end{tabular*}
\end{table}

As shown in Table~\ref{tab:value-runtime}, with resident sender
caches, \textsc{VReparo} retains an average TTFT reduction of 85.62\%
relative to receiver-native inference across the four settings.
This corresponds to approximately $15.2\times$ speedup under full
reuse and $4.5\times$ under DroidSpeak.
Compared with uncorrected reuse, its implementation-level TTFT increment is approximately 34~ms
under full KV reuse and 97~ms with DroidSpeak, while per-token generation
time remains essentially unchanged. The mapping is applied once to
the prefix cache, so it adds no mapping step during subsequent generation.

\section{Conclusion}
\label{sec:conclusion}

In this work, we investigated the safety incompatibility in cross-model KV cache reuse across
different sender--receiver pairs and reuse configurations. Our results show
that cache reuse can change receiver safety in either direction, and that
reused values have a consistently larger effect on safety behavior than
reused keys.
We proposed \textsc{VReparo}, which learns lightweight affine maps
to align reused values with their receiver-native counterparts.
Across the evaluated settings, \textsc{VReparo}
substantially improves safety compatibility while retaining large TTFT
reductions, with trade-offs in benign refusal and response quality.

\FloatBarrier
\subsection*{AI use statement}
\label{sec:ai-use}

We used generative AI tools for literature search, feedback on experimental
design and result interpretation, implementation assistance, and manuscript
preparation. The authors take responsibility for the final manuscript and
accompanying artifacts.

\subsection*{Ethics statement}
\label{sec:ethics}

This work studies safety risks in cross-model KV cache reuse using public
benchmarks. Our experiments do not target deployed third-party services.
We report aggregate results and per-example scores without publishing
harmful model completions.

\subsection*{Reproducibility statement}
\label{sec:reproducibility}

Section~\ref{sec:approach} describes the method. Experimental settings are
provided in Appendix~\ref{app:characterization-settings} for characterization
and Appendix~\ref{app:mitigation-settings} for mitigation.
Appendix~\ref{app:fitting-subsets} provides the sampling procedure and individual
results for the five fitting runs. The code, experiment configurations, and per-example
scores for reproducing the reported tables are available at
\url{https://github.com/Snakinya/VReparo}.

\bibliographystyle{iclr2027_conference}
\bibliography{references}

\appendix
\clearpage
\renewcommand{\topfraction}{.85}
\setcounter{topnumber}{3}
\setcounter{totalnumber}{4}
\makeatletter
\setlength{\@fpsep}{12pt}
\makeatother
\section{\texorpdfstring{Experimental Details for the Characterization Study}{Experimental Details for the Characterization Study}}
\label{app:characterization-settings}

\subsection{\texorpdfstring{Setup and Validity Screening}{Setup and Validity Screening}}

The HarmBench subset comprises textual requests without accompanying
context. We evaluate direct harmful requests without optimizing jailbreak prompts.

We implement cache construction, substitution, and receiver decoding in
PyTorch 2.4.0 with Transformers 4.45.2 and PEFT 0.14.0. All experiments run
on NVIDIA A40 GPUs. Sender and receiver weights and KV caches use FP16
without quantization.
HarmBench uses HarmBench-Llama-2-13b-cls on the first 512 response
tokens. StrongREJECT uses the strongreject-15k-v1 adapter on Gemma-2B with
the last 512 response tokens. Each window uses the evaluator's tokenizer.

RQ1 requires valid outputs under Native and Full KV. RQ2 requires
validity under Native, K-only, V-only, and Full KV. The retained subsets,
and hence their Native ASRs, can therefore differ.

Automatic screening excludes empty, role-only, or export-clipped outputs,
content-free ``the correct answer [is]'' templates, and short prompt-tail echoes
of at most eight words for instruction-initial prompts. Repetition is flagged
when a lowercased four-word sequence occurs at least ten times in an output
of at least 40 words, or a normalized sentence of at least 20 characters and
three words occurs at least four times. A capped-reasoning candidate must
simultaneously reach 512 generated tokens, contain ``the user wants/is
asking/asked,'' ``I need to,'' or ``let me think/see/break/analyze'' within
its first 400 characters, and lack a closing \texttt{</think>} tag.
The cap alone is insufficient.

Existing output-form review annotations marking invalid or uncertain
responses also cause exclusion.
These retrospective, Codex-assisted annotations are not an independent human
audit. Passing the check does not certify semantic validity or benign-task
quality. Reasons and conditions may overlap: Gemma-2-2B RQ2 excludes 38
unique prompts, including 37 V-only failures (36 empty and one repetitive)
and one Full KV capped-reasoning candidate. Qwen3-8B excludes 30 unique
prompts, with 27 Full KV and four V-only failures. Artifacts retain reasons
and overlaps.

\subsection{\texorpdfstring{Detailed Results for Full KV Cache Reuse}{Detailed Results for Full KV Cache Reuse}}
\label{app:rq1-details}
Tables~\ref{tab:rq1-harmbench-detail} and~\ref{tab:rq1-strongreject-detail}
give the full results behind Table~\ref{tab:rq1-full-reuse}, with retained/total
prompt counts $n_{\mathrm{HB}}$ and $n_{\mathrm{SR}}$.
HarmBench $\mathrm{ASR}_{\mathrm{N}}$ and $\mathrm{ASR}_{\mathrm{reuse}}$
denote receiver-native and full-reuse ASRs. StrongREJECT reports average
harmfulness scores for these conditions.
Derived metrics use unrounded scores.

\begin{table}[t]
\caption{Full KV reuse on HarmBench. ASRs and SD/SI/SDV are in \%, and $\Delta S$ is in pp.}
\label{tab:rq1-harmbench-detail}
\setlength{\tabcolsep}{2.7pt}
\renewcommand{\arraystretch}{1.12}
\centering
\begin{tabular*}{\linewidth}{@{\extracolsep{\fill}}lc*{7}{r}@{}}
\toprule
\tablehead{\multirow[c]{2}{*}{Model}}
& \tablehead{\multirow[c]{2}{*}{(Sender, Receiver)}}
& \tablehead{\multirow[c]{2}{*}{$n_{\mathrm{HB}}$}}
& \multicolumn{2}{c}{\textbf{ASR (\%)}}
& \multicolumn{4}{c}{\textbf{Safety incompatibility}} \\
\cmidrule(lr){4-5}\cmidrule(lr){6-9}
& & 
& \tablehead{$\mathrm{ASR}_{\mathrm{N}}$}
& \tablehead{$\mathrm{ASR}_{\mathrm{reuse}}$}
& \tablehead{$\Delta S$}
& \tablehead{SD}
& \tablehead{SI}
& \tablehead{SDV} \\
\midrule
Qwen3-4B & (Base, Instruct) & 158/159 & 10.76 & 20.89 & +10.13 & 13.29 & 3.16 & 16.46 \\
Qwen3-8B & (Base, Instruct) & 132/159 & 6.82 & 4.55 & -2.27 & 0.76 & 3.03 & 3.79 \\
Llama-3.2-3B & (Base, Instruct) & 150/159 & 8.67 & 17.33 & +8.67 & 13.33 & 4.67 & 18.00 \\
Llama-3-8B & (Base, Instruct) & 157/159 & 6.37 & 31.21 & +24.84 & 28.03 & 3.18 & 31.21 \\
Gemma-2-2B & (Base, Instruct) & 158/159 & 1.27 & 21.52 & +20.25 & 20.89 & 0.63 & 21.52 \\
Mistral-7B-v0.3 & (Base, Instruct) & 159/159 & 57.86 & 67.30 & +9.43 & 16.98 & 7.55 & 24.53 \\
\midrule
LoRI-D & (Code, Safety) & 135/159 & 25.93 & 80.74 & +54.81 & 57.78 & 2.96 & 60.74 \\
LoRI-D & (Safety, Code) & 137/159 & 79.56 & 35.04 & -44.53 & 3.65 & 48.18 & 51.82 \\
\bottomrule
\end{tabular*}
\end{table}

\begin{table}[t]
\caption{StrongREJECT scores and incompatibility metrics under full KV reuse (original scale).}
\label{tab:rq1-strongreject-detail}
\setlength{\tabcolsep}{3.5pt}
\renewcommand{\arraystretch}{1.12}
\centering
\begin{tabular*}{\linewidth}{@{\extracolsep{\fill}}lc*{7}{r}@{}}
\toprule
\tablehead{\multirow[c]{2}{*}{Model}}
& \tablehead{\multirow[c]{2}{*}{(Sender, Receiver)}}
& \tablehead{\multirow[c]{2}{*}{$n_{\mathrm{SR}}$}}
& \multicolumn{2}{c}{\textbf{Average score}}
& \multicolumn{4}{c}{\textbf{Safety incompatibility}} \\
\cmidrule(lr){4-5}\cmidrule(lr){6-9}
& & 
& \tablehead{Native}
& \tablehead{Reuse}
& \tablehead{$\Delta S$}
& \tablehead{SD}
& \tablehead{SI}
& \tablehead{SDV} \\
\midrule
Qwen3-4B & (Base, Instruct) & 306/313 & .0614 & .1109 & +.0495 & .0716 & .0221 & .0938 \\
Qwen3-8B & (Base, Instruct) & 296/313 & .0447 & .0186 & -.0260 & .0077 & .0337 & .0414 \\
Llama-3.2-3B & (Base, Instruct) & 259/313 & .0391 & .1173 & +.0781 & .1009 & .0227 & .1236 \\
Llama-3-8B & (Base, Instruct) & 311/313 & .0094 & .2477 & +.2383 & .2387 & .0004 & .2391 \\
Gemma-2-2B & (Base, Instruct) & 313/313 & .0094 & .1658 & +.1564 & .1576 & .0012 & .1589 \\
Mistral-7B-v0.3 & (Base, Instruct) & 311/313 & .4067 & .4949 & +.0882 & .1594 & .0712 & .2306 \\
\midrule
LoRI-D & (Code, Safety) & 287/313 & .1093 & .6741 & +.5648 & .5719 & .0071 & .5790 \\
LoRI-D & (Safety, Code) & 289/313 & .6829 & .2317 & -.4512 & .0187 & .4699 & .4886 \\
\bottomrule
\end{tabular*}
\end{table}

\section{\texorpdfstring{Experimental Details for the Mitigation Evaluation}{Experimental Details for the Mitigation Evaluation}}
\label{app:mitigation-settings}

\subsection{\texorpdfstring{Fitting and Calibration}{Fitting and Calibration}}

\paragraph{DroidSpeak layer selection.}
For DroidSpeak~\citep{liu2026droidspeak}, we profile 136 contiguous blocks
formed from two-layer groups on 50 HotpotQA examples, without observing
safety scores. At each budget, we retain the block with the highest
QA F1 and choose the smallest budget reaching 95\% of native F1.
Both pairs independently select and fix zero-based layers $[0,1]$
for fitting and evaluation.

Each of the five runs uses 32 HotpotQA and 32 HEx-PHI examples.
The four settings use the same examples in each run, with configuration-specific
formatting and separately fitted maps. Appendix~\ref{app:fitting-subsets}
describes sampling and individual results.
We collect paired values at matching positions per layer and KV head,
with $\lambda=1$. For a cache of length $L\le128$, we use
all positions. Otherwise, we take the first 128 positions at stride
$\lfloor L/128\rfloor$, starting at zero and sampling only the reused
shared prefix for DroidSpeak.
The loss sums squared residuals over $N$ positions per head.
$\lambda=1$ is not a mean-loss regularizer. $N$ counts correlated token
positions, not independent texts. Fitting is label-free, but its HEx-PHI inputs are not
benign-only.

We select $\alpha\in\{0.1,0.25,0.5,1\}$ using string-based
refusal on 450 XSTest prompts (250 safe and 200
unsafe)~\citep{rottger2024xstest}. Eligible candidates must increase
unsafe-prompt refusal over uncorrected reuse ($\alpha=0$), while limiting
safe-prompt refusal increases over native to 5~pp overall
and 20~pp per safe subtype. We maximize the unsafe-refusal increase,
breaking ties toward smaller $\alpha$. We select $\alpha$ separately for
each run. Table~\ref{tab:fitting-subsets} gives the selected values.
QA does not determine eligibility. Parameters remain fixed during evaluation, and
the 95\% QA target applies only to DroidSpeak layer profiling.
QA was changed from a development eligibility gate to a reporting
metric. Earlier development artifacts are retained.
\paragraph{Fitting data for timing and additional QA.}
\label{app:reference-fits}
The runtime measurements and additional QA results in
Tables~\ref{tab:value-runtime} and~\ref{tab:value-utility} use maps fitted to the
original 32+32 examples. We selected indices with remainder zero or one modulo
four from each domain's first 64 source examples, giving 32+32 texts.
Their XSTest calibration selects $\alpha=1$ for LoRI-D and $0.25$ for
Llama-3-8B, as detailed in Table~\ref{tab:development-alpha}.
These measurements are not averaged over the five runs.

\begin{table}[t]
\centering
\caption{XSTest development candidates for the maps used
in the timing and additional QA experiments. Unsafe refusal changes
are relative to raw reuse. Safe refusal and maximum safe-subtype changes
are relative to Native (all in pp). Bold $\alpha$ is selected. Each map uses
64 fitting texts. $N$ counts sampled positions per head. These are
string-refusal development results, not independent test sensitivity.}
\label{tab:development-alpha}
\setlength{\tabcolsep}{5pt}
\renewcommand{\arraystretch}{1.12}
\begin{tabular*}{\linewidth}{@{\extracolsep{\fill}}rrrrc@{}}
\toprule
\tablehead{$\alpha$} & \tablehead{Unsafe change} & \tablehead{Safe change} & \tablehead{Max subtype change} & \tablehead{Eligible} \\
\midrule
\multicolumn{5}{@{}l}{\textbf{LoRI-D Full KV, $N=5{,}943$}} \\
0.1 & 2.0 & -21.6 & 0.0 & Yes \\
0.25 & 4.5 & -21.6 & 0.0 & Yes \\
0.5 & 18.0 & -21.2 & 0.0 & Yes \\
\textbf{1.0} & 51.0 & -13.6 & 12.0 & Yes \\
\midrule
\multicolumn{5}{@{}l}{\textbf{LoRI-D DroidSpeak, $N=5{,}815$}} \\
0.1 & 2.0 & -21.6 & 0.0 & Yes \\
0.25 & 8.5 & -21.6 & 0.0 & Yes \\
0.5 & 18.0 & -20.8 & 0.0 & Yes \\
\textbf{1.0} & 58.5 & -16.8 & 0.0 & Yes \\
\midrule
\multicolumn{5}{@{}l}{\textbf{Llama-3-8B Full KV, $N=5{,}335$}} \\
0.1 & 13.5 & 0.0 & 12.0 & Yes \\
\textbf{0.25} & 27.5 & 1.6 & 12.0 & Yes \\
0.5 & 41.5 & 9.6 & 28.0 & No \\
1.0 & 53.5 & 52.4 & 84.0 & No \\
\midrule
\multicolumn{5}{@{}l}{\textbf{Llama-3-8B DroidSpeak, $N=5{,}239$}} \\
0.1 & 11.0 & 0.8 & 12.0 & Yes \\
\textbf{0.25} & 17.0 & 2.8 & 12.0 & Yes \\
0.5 & 24.0 & 12.0 & 24.0 & No \\
1.0 & 32.0 & 54.8 & 72.0 & No \\
\bottomrule
\end{tabular*}
\end{table}

\subsection{\texorpdfstring{Evaluation and Timing Protocol}{Evaluation and Timing Protocol}}

We measure question-answering quality using token-level F1 on HotpotQA,
2WikiMultihopQA~\citep{ho2020multihop}, and MultiFieldQA-en from
LongBench-E~\citep{bai2024longbench}.

\paragraph{Evaluation subsets and refusal scoring.}
The 32 HarmBench development prompts are separate from the fitting
examples. They were selected from an existing 101-prompt subset by cycling
through alphabetically sorted categories, preserving source order within
each category. Excluding their identifiers from the 159-prompt evaluation
pool leaves the 127 prompts used here.
All five runs use the same 248 HotpotQA evaluation examples in all four
settings. The additional QA results in Table~\ref{tab:value-utility} use 149 HotpotQA examples for full
reuse and 248 for DroidSpeak, plus 200 2WikiMultihopQA and 150 MultiFieldQA-en examples.
HotpotQA evaluation subsets exclude fitting, profiling, and calibration
examples by identifier and normalized question/context. XSTest is used
for calibration only.

The 200 OR-Bench prompts include 20 per category. Qwen3-8B judges
responses without condition labels or manual adjudication.
We report unscorable responses separately out of 200.
One response is unscorable for Llama-3-8B full Reuse and for each
of its corrected runs with seeds 0 and 3. The other corrected runs have none.

\paragraph{Generation and safety scoring.}
Generation follows the model-loading setup in
Appendix~\ref{app:characterization-settings}, with greedy decoding and
budgets of 512 tokens for safety/refusal, 32 for HotpotQA and
2WikiMultihopQA, and 64 for MultiFieldQA-en. HarmBench scores the first
512 response tokens using strict single-token yes/no decisions
from HarmBench-Llama-2-13b-cls in BF16.
StrongREJECT scores the last 512 with FP32 evaluator computation
using Gemma-2B merged with strongreject-15k-v1 from BF16-loaded weights.
Window lengths are measured using each evaluator's tokenizer.

\begin{samepage}
Native baselines are evaluated separately for the two reuse configurations.
The full-reuse implementation initializes decoding with a receiver-native
cache, whereas the DroidSpeak implementation uses the receiver's standard
generation routine. We retain each implementation's Native results when
comparing safety and QA quality.
On the safety benchmarks, prompt IDs and texts match, but the saved
Native outputs are not always identical.
We report point estimates.
\par
\end{samepage}

\paragraph{Timing protocol.}
\label{app:mitigation-timing}
We use 50 fixed HotpotQA prompts, five warmups, and five timed repetitions
per condition. Table~\ref{tab:value-runtime} averages the five timed
repetitions per prompt and then the 50 prompts. Each configuration uses one
fixed set of maps and correction strength, as specified in Appendix~\ref{app:reference-fits}.
Native TTFT includes full receiver prefill. Reuse and
\textsc{VReparo} start with resident sender caches and, for DroidSpeak,
intermediate states. TTFT includes mapping where applicable and any receiver
recomputation or suffix processing. The current DroidSpeak implementation
with \textsc{VReparo} includes both the original suffix prefill and an
additional suffix prefill using the corrected cache. Generation time is
measured over 32 forced tokens from prepared caches, including the prompt
boundary step but excluding one-time cache preparation. Sender prefill,
cache transfer, tokenization, and distributed scheduling are excluded.
Table~\ref{tab:timing-lengths} reports prompt lengths. These TTFT
increments include implementation overhead, not just mapping kernels.

\begin{table}[t]
\centering
\caption{Token lengths for 50 timing prompts. Prompt and Suffix
report min/mean/max. Prefix reports the mean. Full KV prefix and suffix
lengths were not recorded separately.}
\label{tab:timing-lengths}
\setlength{\tabcolsep}{4pt}
\renewcommand{\arraystretch}{1.12}
\begin{tabular*}{\linewidth}{@{\extracolsep{\fill}}llrrr@{}}
\toprule
\tablehead{Model pair} & \tablehead{Reuse} & \tablehead{Prompt} & \tablehead{Prefix} & \tablehead{Suffix} \\
\midrule
LoRI-D & Full KV & 1,833/7,262.20/7,530 & --- & --- \\
 & DroidSpeak & 1,833/7,261.30/7,529 & 7,210.04 & 42/51.26/80 \\
Llama-3-8B & Full KV & 1,814/7,243.20/7,511 & --- & --- \\
 & DroidSpeak & 1,814/7,242.30/7,510 & 7,191.04 & 42/51.26/80 \\
\bottomrule
\end{tabular*}
\end{table}

\subsection{\texorpdfstring{QA Quality}{QA Quality}}
\label{app:mitigation-quality}

\begin{table}[t]
\caption{QA token F1 ($\times100$). Fitting data are specified in Appendix~\ref{app:reference-fits}. Native and Reuse are shown separately for each configuration. HotpotQA uses 149 examples for full reuse and 248 for DroidSpeak. 2Wiki and MultiField denote 2WikiMultihopQA and MultiFieldQA-en. Bold marks the best F1 per row and dataset.}
\label{tab:value-utility}
\setlength{\tabcolsep}{2.4pt}
\renewcommand{\arraystretch}{1.12}
\centering
\begin{tabular*}{\linewidth}{@{\extracolsep{\fill}}l*{9}{r}@{}}
\toprule
\tablehead{\multirow[c]{2}{*}{Config.}}
& \multicolumn{3}{c}{\textbf{HotpotQA}}
& \multicolumn{3}{c}{\textbf{2Wiki}}
& \multicolumn{3}{c}{\textbf{MultiField}} \\
\cmidrule(lr){2-4}\cmidrule(lr){5-7}\cmidrule(l){8-10}
&
\tablehead{Native}
& \tablehead{Reuse}
& \tablehead{\textsc{VReparo}}
& \tablehead{Native}
& \tablehead{Reuse}
& \tablehead{\textsc{VReparo}}
& \tablehead{Native}
& \tablehead{Reuse}
& \tablehead{\textsc{VReparo}} \\
\midrule
\multicolumn{10}{@{}l}{\textbf{LoRI-D}} \\
Full KV & \tableemph{28.90} & 22.48 & 21.13
& \tableemph{27.85} & 25.19 & 22.03
& \tableemph{50.71} & 48.71 & 46.42 \\
DroidSpeak & 27.36 & 27.90 & \tableemph{28.92}
& \tableemph{27.85} & 27.40 & 26.99
& 50.75 & \tableemph{51.38} & 49.55 \\
\midrule
\multicolumn{10}{@{}l}{\textbf{Llama-3-8B}} \\
Full KV & \tableemph{49.64} & 9.64 & 11.54
& \tableemph{34.27} & 10.60 & 11.59
& \tableemph{44.45} & 13.28 & 13.50 \\
DroidSpeak & \tableemph{47.46} & 43.43 & 43.36
& 34.27 & \tableemph{40.64} & 40.52
& \tableemph{44.61} & 41.94 & 41.67 \\
\bottomrule
\end{tabular*}
\end{table}

Table~\ref{tab:value-utility} distinguishes the incremental effect of
\textsc{VReparo} from the quality loss already introduced by reuse.
Under full KV reuse, LoRI-D loses F1 on all three datasets, with the
largest relative drop of 12.52\% on 2WikiMultihopQA compared with uncorrected reuse. Llama-3-8B improves
over uncorrected reuse but remains below native quality. Changes relative
to uncorrected DroidSpeak are smaller: LoRI-D improves on HotpotQA and
declines on the other two datasets, while Llama-3-8B declines slightly
on all three. The 95\% profiling target does not guarantee evaluation
quality: Llama-3-8B DroidSpeak retains only 91.52\% of native HotpotQA
F1 before \textsc{VReparo}.

\begin{samepage}
\subsection{\texorpdfstring{Safety Scores and Benign Refusal}{Safety Scores and Benign Refusal}}
\label{app:mitigation-absolute}
Table~\ref{tab:value-absolute} complements the Native-referenced metrics
in Table~\ref{tab:value-safety} with HarmBench ASR, average StrongREJECT
scores, and OR-Bench full refusal rates. All metrics use the same evaluation prompts as the main table.
Paired metrics use unrounded scores and may differ from differences
of displayed values.
\par
\end{samepage}

\begin{table}[t]
\centering
\caption{Safety scores and benign refusal.
HarmBench ASR and OR-Bench full Refusal are in \%.
Mean is the average StrongREJECT score (0--1).
\textsc{VReparo} results are averaged over five runs.}
\label{tab:value-absolute}
\setlength{\tabcolsep}{5pt}
\renewcommand{\arraystretch}{1.12}
\begin{tabular*}{\linewidth}{@{\extracolsep{\fill}}lc*{6}{r}@{}}
\toprule
\tablehead{\multirow[c]{2}{*}{Model}}
& \tablehead{\multirow[c]{2}{*}{Condition}}
& \multicolumn{3}{c}{\textbf{Full KV reuse}}
& \multicolumn{3}{c}{\textbf{DroidSpeak}} \\
\cmidrule(lr){3-5}\cmidrule(l){6-8}
& &
\tablehead{ASR}
& \tablehead{Mean}
& \tablehead{Refusal}
& \tablehead{ASR}
& \tablehead{Mean}
& \tablehead{Refusal} \\
\midrule
LoRI-D & Native & 29.13 & 0.1219 & 11.0 & 29.13 & 0.1213 & 11.0 \\
 & Reuse & 79.53 & 0.6431 & 4.0 & 82.68 & 0.6353 & 4.5 \\
 & \tableemph{\textsc{VReparo}} & 64.09 & 0.3966 & 12.6 & 60.47 & 0.3916 & 10.8 \\
\midrule
Llama-3-8B & Native & 7.09 & 0.0107 & 10.0 & 7.09 & 0.0108 & 9.5 \\
 & Reuse & 34.65 & 0.2505 & 12.5 & 32.28 & 0.1569 & 8.5 \\
 & \tableemph{\textsc{VReparo}} & 14.65 & 0.0373 & 25.3 & 17.64 & 0.0398 & 17.1 \\
\bottomrule
\end{tabular*}
\end{table}

\FloatBarrier
\subsection{\texorpdfstring{Data Sampling and Detailed Results}{Data Sampling and Detailed Results}}
\label{app:fitting-subsets}

We sample 32 HotpotQA and 32 HEx-PHI examples for each run using
seeds 0--4. After excluding profiling and evaluation
overlaps by identifier or normalized text, the candidate pools contain 143 QA
and 330 safety examples. Sampling balances QA examples across three length
groups and safety examples across all 11 HEx-PHI categories.
Examples may overlap across runs. Each run uses the same examples across
the four settings, with maps fitted separately
and $\alpha$ selected on XSTest using the rule in Appendix~\ref{app:mitigation-settings}.
Other fitting hyperparameters and the DroidSpeak layer block remain fixed.

\begin{table}[t]
\centering
\caption{Results for each sampling seed. HB is SDV
on 127 HarmBench prompts (\%). SR is SDV on 313 StrongREJECT prompts.
F1 is HotpotQA token F1 ($\times100$) on 248 prompts, and Refusal is the full-refusal
rate on 200 OR-Bench prompts (\%).}
\label{tab:fitting-subsets}
\small
\setlength{\tabcolsep}{3pt}
\renewcommand{\arraystretch}{1.12}
\begin{tabular*}{\linewidth}{@{\extracolsep{\fill}}llrrrrrr@{}}
\toprule
\tablehead{Model} & \tablehead{Config.} & \tablehead{Seed} & \tablehead{$\alpha$} & \tablehead{HB} & \tablehead{SR} & \tablehead{F1} & \tablehead{Refusal} \\
\midrule
LoRI-D & Full KV & 0 & 1 & 44.09 & 0.3149 & 20.05 & 12.5 \\
 & & 1 & 1 & 40.94 & 0.3105 & 20.37 & 13.0 \\
 & & 2 & 1 & 44.09 & 0.3156 & 20.25 & 13.0 \\
 & & 3 & 1 & 42.52 & 0.3094 & 20.59 & 12.5 \\
 & & 4 & 1 & 42.52 & 0.3163 & 19.94 & 12.0 \\
\addlinespace[4pt]
 & DroidSpeak & 0 & 1 & 40.16 & 0.3121 & 28.84 & 10.5 \\
 & & 1 & 1 & 36.22 & 0.2833 & 28.21 & 10.5 \\
 & & 2 & 1 & 38.58 & 0.3103 & 28.61 & 10.5 \\
 & & 3 & 1 & 38.58 & 0.2917 & 28.23 & 11.0 \\
 & & 4 & 1 & 39.37 & 0.2969 & 28.63 & 11.5 \\
\midrule
Llama-3-8B & Full KV & 0 & 0.5 & 12.60 & 0.0308 & 10.15 & 25.0 \\
 & & 1 & 0.5 & 13.39 & 0.0331 & 10.09 & 25.5 \\
 & & 2 & 0.5 & 13.39 & 0.0354 & 11.43 & 25.0 \\
 & & 3 & 0.5 & 12.60 & 0.0313 & 11.06 & 24.5 \\
 & & 4 & 0.5 & 11.02 & 0.0273 & 10.03 & 26.5 \\
\addlinespace[4pt]
 & DroidSpeak & 0 & 0.5 & 11.02 & 0.0248 & 43.28 & 19.5 \\
 & & 1 & 0.5 & 11.02 & 0.0231 & 43.46 & 19.0 \\
 & & 2 & 0.25 & 16.54 & 0.0581 & 43.32 & 13.5 \\
 & & 3 & 0.25 & 17.32 & 0.0548 & 42.76 & 13.0 \\
 & & 4 & 0.5 & 11.02 & 0.0202 & 44.28 & 20.5 \\
\bottomrule
\end{tabular*}
\end{table}

SDV is lower than uncorrected reuse in all five runs for all four
settings on both benchmarks. All five runs select $\alpha=1$ for LoRI-D
and $0.5$ for Llama-3-8B full reuse. Llama-3-8B DroidSpeak selects $0.5$
in three runs and $0.25$ in two.

\end{document}